\documentclass[conference]{IEEEtran}
\IEEEoverridecommandlockouts
\usepackage{cite}
\usepackage{amsmath,amssymb,amsfonts}
\usepackage{algorithmic}
\usepackage{graphicx}
\usepackage{textcomp}
\usepackage{xcolor}
\usepackage{graphicx}
\usepackage{subfigure}
\def\BibTeX{{\rm B\kern-.05em{\sc i\kern-.025em b}\kern-.08em
    T\kern-.1667em\lower.7ex\hbox{E}\kern-.125emX}}

\usepackage{booktabs}
\usepackage{comment}
\usepackage{amsmath}

\usepackage{algorithm}
\usepackage{algorithmic,esint}
\usepackage{enumitem}
\usepackage{graphicx}
\usepackage{newfloat}
\usepackage{listings}
\usepackage{subfig}
\usepackage{adjustbox}
\usepackage{wrapfig}

\newtheorem{assumption}{Assumption}

\newcommand{\model}{\textbf{DyG-HAP} }
\newcommand{\modelnosp}{\textbf{DyG-HAP}}
\newcommand{\red}[1]{\textcolor{black}{#1}}
\newcommand{\ms}[2]{{#1}\scriptsize{$\pm$#2}}
\newcommand{\msone}[2]{\bf {#1}\scriptsize{$\pm$#2}}
\newcommand{\mstwo}[2]{\underline{{#1}\scriptsize{$\pm$#2}}}
\newcommand{\ie}{\textit{ i.e.}}
\newcommand{\eg}{\textit{ e.g.}}
\newcommand{\etal}{\textit{ et al.}}
\usepackage{tabularx}

\begin{document}

\title{Out-of-Distribution Generalized Dynamic Graph Neural Network for Human Albumin Prediction\\
}

\author{
\IEEEauthorblockN{Zeyang Zhang\textsuperscript{*}\thanks{${}^*$Equal contributions}}
\IEEEauthorblockA{\textit{Computer Science and Technology} \\
\textit{Tsinghua Univerisity}\\
Beijing, China \\
zy-zhang20@mails.tsinghua.edu.cn}
\\

\IEEEauthorblockN{Ning Lin}
\IEEEauthorblockA{\textit{  College of Medicine  } \\
\textit{Southwest Jiaotong University}\\
Sichuan, China \\
helenmedic@yeah.net}

\and

\IEEEauthorblockN{Xingwang Li\textsuperscript{*}}
\IEEEauthorblockA{\textit{Computing and Artificial Intelligence} \\
\textit{Southwest Jiaotong University}\\
Sichuan, China \\
xingwangli98@163.com}
\\
\IEEEauthorblockN{Xueling Zhu\textsuperscript{$\dagger$}}
\IEEEauthorblockA{\textit{Xiangya School of Medicine} \\
\textit{Central South University}\\
Hunan, China \\
xuercan@163.com}

\and
\IEEEauthorblockN{Fei Teng}
\IEEEauthorblockA{\textit{Computing and Artificial Intelligence} \\
\textit{Southwest Jiaotong University}\\
Sichuan, China \\
fteng@swjtu.edu.cn}
\\
\IEEEauthorblockN{Xin Wang\textsuperscript{$\dagger$}\thanks{${}^\dagger$Corresponding authors}, Wenwu Zhu}
\IEEEauthorblockA{\textit{Computer Science and Technology} \\
\textit{Tsinghua Univerisity}\\
Beijing, China \\
\{xin\_wang, wwzhu\}@tsinghua.edu.cn}

}
\maketitle

\begin{abstract}
Human albumin is essential for indicating the body's overall health. Accurately predicting plasma albumin levels and determining appropriate doses are urgent clinical challenges, particularly in critically ill patients, to maintain optimal blood levels. However, human albumin prediction is non-trivial that has to leverage the dynamics of biochemical markers as well as the experience of treating patients. Moreover, the problem of distribution shift is often encountered in real clinical data,  which may lead to a decline in the model prediction performance and reduce the reliability of the model's application. In this paper, we propose a framework named Out-of-Distribution Generalized Dynamic Graph Neural Network for Human Albumin Prediction (\modelnosp), which is able to provide accurate albumin predictions for Intensity Care Unit (ICU) patients during hospitalization. We first model human albumin prediction as a dynamic graph regression problem to model the dynamics and patient relationship. Then, we propose a disentangled dynamic graph attention mechanism to capture and disentangle the patterns whose relationship to labels under distribution shifts is invariant and variant respectively. Last, we propose an invariant dynamic graph regression method to encourage the model to rely on invariant patterns to make predictions. Moreover, we propose a dataset named Albumin level testing and nutritional dosing data for Intensive Care (ANIC) for evaluation. Extensive experiments demonstrate the superiority of our method compared to several baseline methods in human albumin prediction. 
\end{abstract}


\section{Introduction}

Human albumin plays a critical role in maintaining blood osmolarity, providing protection, facilitating transport, regulating functions, and combating inflammation. The levels of serum albumin are indicative of the body's overall health\cite{di2019long}. Furthermore, hypoalbuminemia can manifest as a symptom of nephrotic syndrome, which is characterized by kidney damage and partial protein loss\cite{hashim2020response}. Consequently, this condition results in elevated protein levels in the urine and decreased serum albumin levels. Therefore, accurately predicting plasma albumin levels and determining albumin doses become urgent clinical tasks to maintaining optimal blood levels in critically ill patients.

In recent years, there have been many related works emerging to forecast the patients' biochemical marker concentrations, including albumin. For instance,  Kaji\etal~\cite{kaji2019attention} employ the long short-term memory network (LSTM) to predict clinical events, including sepsis, myocardial ischemia, and antibiotic vancomycin. Cheng\etal~\cite{DBLP:conf/kdd/LiuWHX15} adopt a convolutional neural network (CNN) to construct a temporal matrix of medical codes from multiple patients, focusing on predicting congestive heart failure diseases. Choi\etal~\cite{DBLP:conf/mlhc/ChoiBSSS16} propose a gated recurrent unit (GRU) that predicts the timing of a patient's next visit based on diagnostic codes. These methods have demonstrated competitive results in capturing the dynamics of several biochemical markers.

However, the existing methods commonly assume that the training and testing data are independently sampled from the same distribution, which may not hold in real-world data due to the uncontrollable distribution shifts between training and deployment. 
For example, patients from various regions or periods may differ greatly in lifestyle, environmental factors, dietary habits, {\it etc.}, and thus have different dynamics of biochemical markers. The model trained on the patients from specific regions and periods may exploit variant patterns, whose relationship between the labels is variant under distribution shifts, to make predictions, and struggle to capture the specific factors influencing albumin levels when tested on patients from other regions or periods. In this case, the model will suffer severe performance deterioration from the out-of-distribution samples, and make unreliable albumin predictions. 

In this paper, we study the problem of human albumin prediction, which faces three critical challenges: 1) how to model the dynamics of the albumin levels for each patient, 2) how to leverage the similarity between patients to improve the prediction accuracy, 3) how to handle distribution shifts that naturally exist in the real-world data. To this end, we propose a framework named Out-of-Distribution Generalized \underline{Dy}namic \underline Graph Neural Network for \underline Human \underline Albumin \underline Prediction (\modelnosp), which is able to provide accurate albumin predictions for the patients in the Intensity Care Unit (ICU) during hospitalization. Specifically, we first model the human albumin prediction problem as a dynamic graph regression problem to simultaneously consider the influences of the dynamic patient relationship and attributes on the albumin levels. Second, we propose a disentangled dynamic graph attention mechanism to capture the high-order graph dynamics and disentangle the patterns whose relationship to labels under distribution shifts is invariant and variant respectively.  Last, we propose an invariant dynamic graph regression method that minimizes the variance of the model predictions under exposure to different variant patterns. This encourages the model to rely on the invariant patterns to make predictions and thus handle distribution shifts. Moreover, we propose a dataset named Albumin level testing and nutritional dosing data for Intensive Care (ANIC), which is real patient data collected from ICU that contains various features like demographic characteristics, nutritional support, biochemical markers, {\it etc.} Extensive experiments show that our method achieves state-of-the-art performance over several baseline methods in terms of human albumin prediction. To summarize, we make the following contributions:

\begin{itemize}[leftmargin=0.5cm]
    \item We propose to study human albumin prediction with dynamic graph neural networks, to the best of our knowledge, for the first time.
    
    \item We propose a framework named Out-of-Distribution Generalized Dynamic Graph Neural Network for Human Albumin Prediction (\modelnosp), which is able to provide accurate albumin predictions for ICU patients during hospitalization.

    \item  We propose a dataset named Albumin level testing and nutritional dosing data for Intensive Care (ANIC), which is real patient data collected from ICU that contains various dynamic features and structures.
    
    \item We conduct extensive experiments to demonstrate the superior performance of our method compared to state-of-the-art baselines for human albumin predictions.
    
\end{itemize}

\begin{table*}[htbp]
\centering
	\caption{The descriptions and statistics of the biochemical markers in the ANIC Dataset}
	\label{tab:freq}
        \renewcommand\arraystretch{1.4}
        \setlength{\tabcolsep}{1.2mm}
	\begin{tabular}{cccccc}
		\toprule
		Attribute&Units&Average&Standard Deviation&Range&Description\\
		\midrule
            Albumin (ALB) &g/L & 36.9&5.5&16.2 $\sim$ 56.2&Protein level in blood plasma\\
            Indirect Bilirubin& umol/L& 12.6&13.0&0.2 $\sim$ 131.6&Unconjugated bilirubin in blood\\
            Total Bilirubin (TB) &umol/L & 25.4&49.2&3.2 $\sim$ 615.1& Sum of direct and indirect bilirubin \\
		Direct Bilirubin &umol/L & 12.7&36.7&0.8 $\sim$ 495.6&Conjugated bilirubin in blood\\
            Hemoglobin concentration (Hb) &g/L &106.5&23.2&46.0 $\sim$ 195.0&Amount of hemoglobin in blood\\
            Alanine Aminotransferase (ALT) &u/L & 85.4&296.9&1.6 $\sim$ 4223.0&Liver enzyme marker\\
            Aspartate Aminotransferase (AST) & u/L& 102.0&434.9&4.4 $\sim$ 10268.9
&Liver and heart enzyme marker \\
            Mean RBC hemoglobin content (MCH)&pg/L & 30.0&2.5&19.4 $\sim$ 38.3
&Mean hemoglobin in red blood cells\\
	    Mean RBC hemoglobin concentration (MCHC)&g/L& 324.4&13.6&275.0 $\sim$ 405.0& Mean hemoglobin concentration in red blood cells\\
		\bottomrule
	\end{tabular}
\end{table*} 

\section{Materials}
In this section, we introduce a dataset named Albumin level testing and nutritional dosing data for Intensive Care (ANIC).

\subsection{Dataset Construction}

In this paper, we use the nutrition data of critically ill patients provided by a tertiary hospital in Sichuan Province, which contained basic information, laboratory index data, nutrition and medication use during hospitalization for 5727 ICU patients between 2014 and 2020, with 31106 records. Specifically, patient demographic information is used for admission registration and contains fields such as patient ID, gender, age, diagnosis, and record time. Nutritional support data records patients' medication administration records from admission to discharge (or death), including fields such as patient ID, medical advice, dosage, recording time, the nurse starts execution time and end execution time. The examination data reflects the patient's physical health status and provides a reference basis for subsequent clinical nutrition support, including patient ID, examination index and value. 

Moreover, during the construction of the dataset, we consider the time of patients' admission to the ICU and the medical orders they received as follows:

\begin{itemize}[leftmargin=0.5cm]

\item Initially, for patients with multiple admissions to the ICU, their medical records would contain several transfer records, leading to ambiguous time delineation. To address this issue, we opted to consider the date of the patient’s first admission to the ICU as the representative temporal information in the medical record.

\item Subsequently, concerning medical orders, there are instances where a patient's record indicates the use of different nutritional formulations within a single day—either singularly or in combination. This irregularity largely stems from some physicians' non-standard medical orders, resulting in prescribed daily nutritional dosages that deviate from conventional clinical practices. To remedy this, after consulting and agreeing upon standards with the attending physicians, we meticulously reviewed and rectified the medical orders documented in the dataset, ensuring accuracy and conformity to established clinical norms.

\end{itemize}
\subsection{ Dataset Preprocessing}

In this study, patients aged from 18 to 65 were selected based on the guidance of doctors, where we exclude children due to their incomplete growth and underdeveloped renal function, and elderly people due to their unique physiological characteristics in clinical practice.
Additionally, patients who had albumin test values for 12 consecutive days were screened for inclusion in the study cohort. After consultation with clinicians, nine biochemical markers related to albumin were selected for analysis, and their descriptions and statistics are presented in Table~\ref{tab:freq}. We also utilize categorical features including demographic characteristics and nutrition support, and their descriptions are shown in Table~\ref{tab:cat}.

\begin{table}[]
\centering
\caption{The descriptions of the categorical features in the ANIC dataset, including demographic characteristics and nutritional support. `C' denotes the number of categories. }
\label{tab:cat}

\begin{tabularx}{0.48\textwidth}{lrX}
\toprule
Attribute & C & Category Descriptions                                    \\ \midrule
Age       & 2                                 & Age $\leq$ 45, Age $>$ 45                                    \\
Sex       & 2                                 & Male, Female                                             \\
Disease   & 3                                 & Trauma , Surgery , Internal         \\
EN        & 2                                 & Use enteral nutrition support                    \\
PN        & 2                                 & Use parenteral nutrition support                  \\
PN\_EN     & 2                                 & Use both parenteral and enteral nutrition support \\
ALB       & 2                                 & Use serum albumin                                 \\
EN\_ALB    & 2                                 & Use enteral serum albumin                         \\
PN\_ALB    & 2                                 & Use parenteral serum albumin                      \\
EN\_PN\_ALB & 2                                 & Use both parenteral and enteral serum albumin    \\ \bottomrule
\end{tabularx}

\end{table}

\begin{figure*}
\centering
\includegraphics[width=0.95\textwidth]{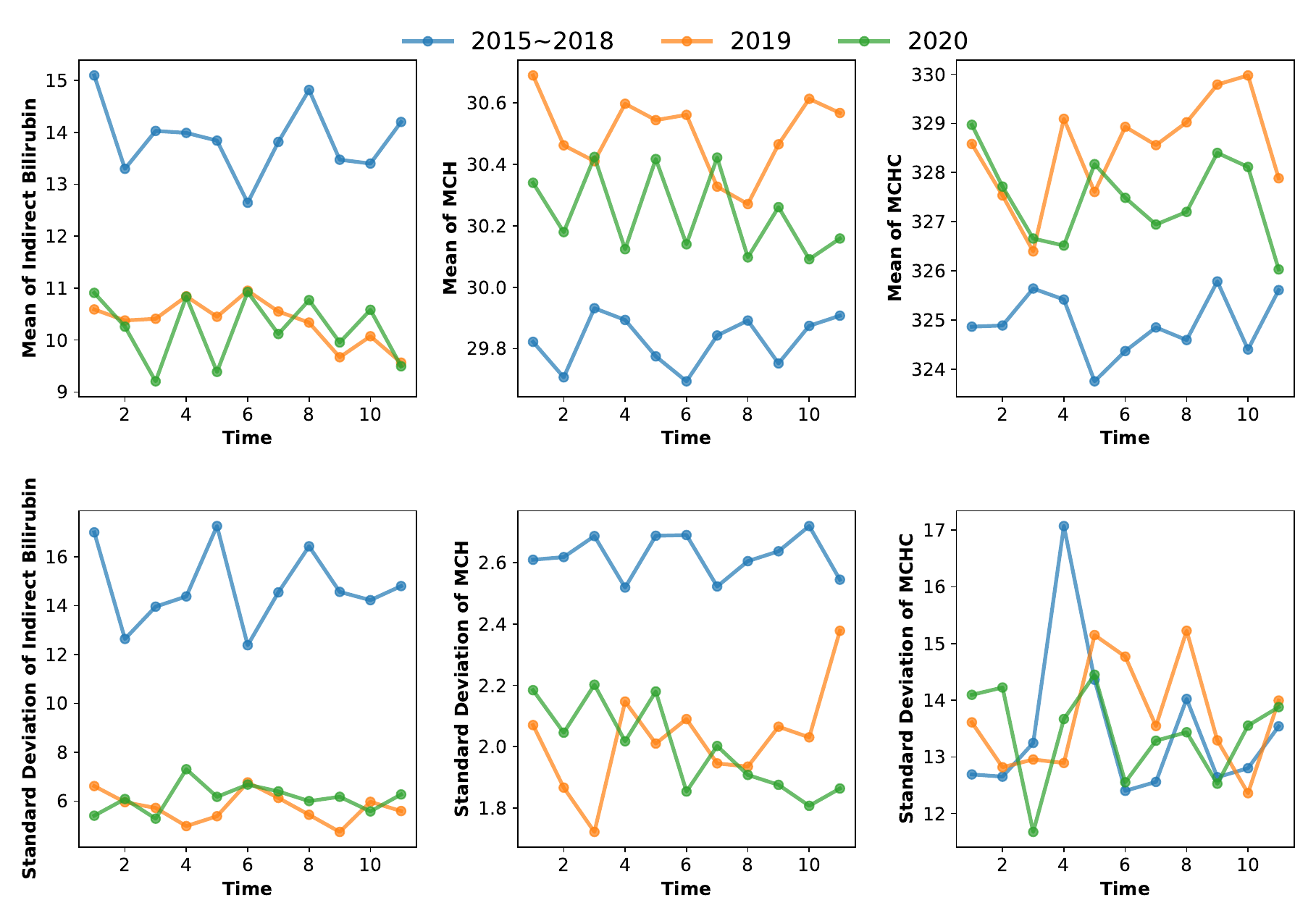}
\caption{Changes in the distribution of biochemical markers over time.}
\label{fig:dis}
\end{figure*}

\subsection{Distributional shifts in the ANIC dataset}
We analyze the ANIC dataset to illustrate the problem of distributional shift in real clinical data. Specifically, we screen all patient test data for 2015, 2019, and 2020 and use eleven days as a treatment cycle for patients to observe health status change. In addition, we select three detection indicators, namely `Indirect bilirubin', `MCH', `MCHC', and calculate the mean and standard deviation of the three indicators. 

The reasons for selecting these three indicators are as follows:
Indirect bilirubin represents the unbound bilirubin present in the bloodstream. An elevation in its concentration could be attributed to the excessive lysis of red blood cells or liver dysfunction, leading to an aberration in bilirubin metabolism. This phenomenon could, in turn, impinge upon the synthesis of albumin, given that the liver serves as the primary site for albumin production. In addition, MCH and MCHC are indicative of the hemoglobin content within red blood cells. A reduction in these indices could signify the onset of anemia or malnutrition in patients, influencing the levels of albumin consequently.

Hence, there exists a salient correlation amongst indirect bilirubin, MCH, MCHC, and albumin levels, collectively delineating the nutritional and organ functional status of patients. The interrelationship underscores their combined utility in offering a comprehensive insight into the physiological well-being of individuals, facilitating nuanced clinical evaluations and interventions.

As shown in Figure~\ref{fig:dis}, the distributions of indicators are constantly shifted as the year progresses. Compared with the detection data in 2015, the data distribution in 2019 and 2020 is closer, indicating that the degree of detection indicator shift is positively correlated with time. In addition, the trend of the detection data is different during six years, with a significant oscillation of the indicator curve in 2020. It suggests that the health status of most patients is unstable, which may be related to the patient's disease state, treatment regimen, lifestyle, or other factors. We could infer that the patient experienced a large physiological or pathological change in 2020. Notably, the COVID-19 pandemic in 2020 may have impacted patients' health status. Therefore, the apparent oscillation of the indicator curve in 2020 may be related to the pandemic, and patients may have been affected by the pandemic, e.g., increased risk of infection and strained healthcare resources, which led to the instability of the indicator changes.

\section{Problem Formulation}
In this section, we formulate the problem of human albumin prediction. For a patient in Intensive Care Unit (ICU), our target is to predict the patient's albumin in the future during hospitalization, which can provide some assistance and guidance for subsequent treatments such as nutritional support. The simplest formulation is {\it uni-variate time-series problem},\ie,
\begin{equation}
\hat y_{n,t} = f(y_{n,1},y_{n,2},\dots,y_{n,t-1}),
\end{equation}
where $y_{n,t} \in \mathbb{R}$ is the albumin of the $n$-th patient at time $t$, $f(\cdot)$ is the prediction function to be learned, and $\hat y_{n,t}$ is the predicted albumin. This formulation exploits the patient's albumin history to predict the albumin at the next time. However, other physiological factors are closely related to human albumin,\eg, Bilirubin, Hb, {\it etc}. In comparison, {\it multi-variate time-series problem} is a more complex yet powerful formulation that can consider the influence of other factors on the patients' human albumin changes,\ie,
\begin{equation}
\hat y_{n,t} = f(\mathbf{x}_{n,1},\mathbf{x}_{n,2},\dots,\mathbf{x}_{n,t-1}),
\end{equation}
where $\mathbf{x}_{n,t} \in \mathbb{R}^{d}$ is the features of the $n$-th patient at time $t$, $d$ denotes the dimensionality of the features, and features can include physiological characteristics, the nutritional support, {\it etc.} As the patients similar in physiological characteristics may have similar albumin dynamics, in this paper, we further consider the relationship between the patients to improve the accuracy of human albumin prediction, by formulating the {\it dynamic graph prediction problem}. Consider a graph $\mathcal{G}$ with the node set $\mathcal{V}$ and the edge set $\mathcal{E}$, where the nodes denote the patients, and the edges denote the relationship between the patients. A dynamic graph can be defined as 
$\mathcal{G}=(\{\mathcal{G}_{t}\}_{t=1}^{T})$, 
where $T$ is the number of time, $\mathcal{G}_t=(\mathcal{V}_t,\mathcal{E}_t)$ is the graph snapshot at time $t$,  $\mathcal{V}=\bigcup_{t=1}^{T} \mathcal{V}_{t}$, $\mathcal{E}=\bigcup_{t=1}^{T} \mathcal{E}_{t}$. For simplicity, a graph slice is also denoted as $\mathcal{G}^t=(\mathbf{X}_t,\mathbf{A}_t)$, which includes node features and adjacency matrix at time $t$. The element in the adjacency matrix $\mathbf{A}_{ij}$ denotes the relationship between the $i$-th patient and the $j$-th patient. Then the prediction task is 
\begin{equation}
\hat{\mathbf{Y}}_{t} = f(\mathcal{G}_{1},\mathcal{G}_{2},\dots,\mathcal{G}_{t-1}),
\end{equation}
where $\mathbf{Y}_{t} \in \mathbb{R}^{N}$ is the albumin of $N$ patients at time $t$. In classical machine learning literature, it is commonly assumed that the data is identically independently distributed, and the model trained with empirical risk minimization in training data is expected to generalize in testing data. However, in real-world applications, there usually exists a gap between the training and testing distributions,\ie,
\begin{equation}
\small
p_{\text{train}}(\mathbf{Y}_t, \mathbf{G}_1,\mathbf{G}_2,\dots,\mathbf{G}_{t-1}) \neq p_{\text{test}}(\mathbf{Y}_t, \mathbf{G}_1,\mathbf{G}_2,\dots,\mathbf{G}_{t-1}),
\end{equation}
which causes an {\it out-of-distribution (OOD) generalization problem}. For example, the distributions of biochemical markers and albumin dynamics change in different years, and the model trained with data before 2019 may have deteriorated prediction performance for the data after 2019. Therefore, it is critical to improve the out-of-distribution generalization abilities of the prediction models. For the OOD generalization literature~\cite{zhang2022dynamic,wu2022discovering, gagnon2022woods,zhang2023outofdistribution,zhang2023spectral,arjovsky2019invariant,chang2020invariant,ahuja2020invariant}, we make the following assumption.
\begin{assumption} For a given task, there exists a predictor $f(\cdot)$, for samples ($\mathcal{G}_{1:t}$,$\mathbf{Y}_{t}$) from any distribution, there exists an invariant pattern $P^I_t$ and a variant pattern $P^V_t$  such that $\mathbf{Y}_t=f(P_t^I)+\epsilon$ and $P_t^I = \mathcal{G}_{1:t} \backslash P^V$, i.e., $\mathbf{Y}_t \perp \mathbf{P}^V_t \mid \mathbf{P}^I_t$. 
\label{assum}
\end{assumption}

This assumption shows that there exist patterns,\ie, a part of patients' evolving features and structures, whose relationship to the patients' albumins is invariant across distributions. To improve the OOD generalization ability, the model has to exploit the invariant patterns to make predictions.

\section{Method}

\begin{figure*}[htbp]
\centering
\includegraphics[width = 0.85\textwidth]{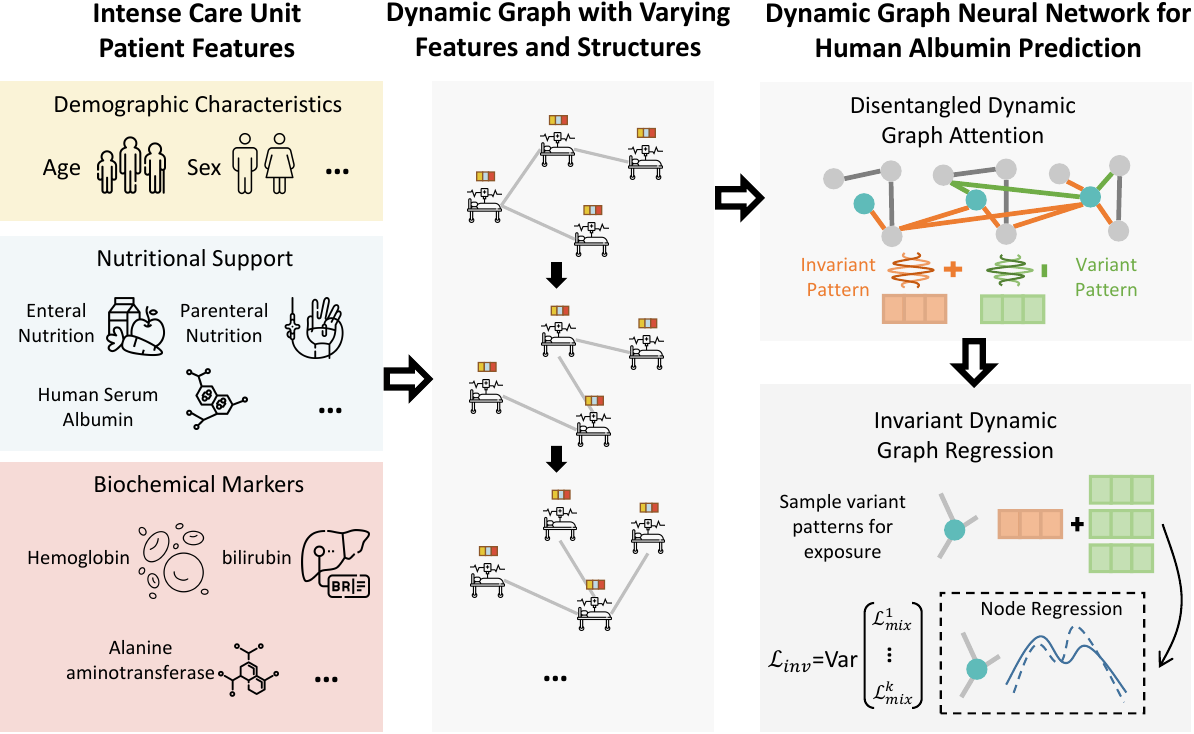}
\caption{The framework of our proposed Out-of-Distribution Generalized Dynamic Graph Neural Network for Human Albumin Prediction (\modelnosp). (Left) We collect the ANIC dataset that has various patient features, including demographic characteristics, nutritional support, biochemical markers, {\it etc}. (Middle) We model the human albumin problem as dynamic graph regression and construct a dynamic graph with varying features and structures to simultaneously consider the dynamic patient relationship and attributes. (Right) We propose a model composed of a disentangled dynamic graph attention that captures and disentangles the invariant and variant patterns, and an invariant dynamic graph regression method to encourage the model to rely on invariant patterns to make predictions.}
\label{fig:framework}
\end{figure*}

In this section, we introduce our Out-of-Distribution Generalized Dynamic Graph Neural Network for Human Albumin Prediction (\modelnosp). 
First, we propose a disentangled dynamic graph attention network to extract the invariant and variant patterns behind the patients' evolving features and relationships. 
Then we propose an invariant dynamic graph regression method to encourage the model to rely on invariant patterns whose relationship to albumins is invariant across distributions to make predictions.

\subsection{Disentangled Dynamic Graph Attention Network}
Since the features and the relationships of the patients are evolving through time, we first design a temporal graph attention mechanism to simultaneously aggregate structural and temporal information in the dynamic graph. Denote the neighborhood of the node $u$ at time $t$ as $\mathcal{N}_t^u = \{v: (u,v)\in \mathcal{E}_t \}$. For a node $u$ at time $t$ and its neighbors $v \in \mathcal{N}_{t'}^u, \forall t'\leq t$, we calculate the Query-Key-Value vectors as
\begin{equation}
\begin{aligned}
\mathbf{q}_t(u)&=\mathbf{W}_q\Bigl(\mathbf{h}_t(u) + \text{TE}(t)\Bigr),\\
\mathbf{k}_{t'}(v)&=\mathbf{W}_k\Bigl(\mathbf{h}_{t'}(v) + \text{TE}(t')\Bigr),\\
\mathbf{v}_{t'}(v)&=\mathbf{W}_v\Bigl(\mathbf{h}_{t'}(v) + \text{TE}(t')\Bigr),\\
\end{aligned}
\end{equation}
where $\mathbf{h}_t^u$ denotes the representation of node $u$ at the time $t$, $\mathbf{q}$, $\mathbf{k}$, $\mathbf{v}$ represents the query, key and value vector, respectively, and the bias term is omitted for brevity. Following~\cite{xu2020inductive}, we adopt a temporal encoding technique to consider the temporal information inherently,\ie, 
\begin{equation}
\text{TE}(t) = [\text{sin}(\omega_1 t),\text{sin}(\omega_2 t), \dots, \text{sin}(\omega_d t))],
\end{equation}
where $d$ denotes the dimensionality of the node embeddings. Note that $\text{TE}(\cdot)$ can be viewed as a kernel method that can learn the relative temporal information through multiplication,\ie, $\text{TE}(t-t') = \text{TE}(t) \cdot \text{TE}(t')$. In this way, the temporal information is naturally incorporated in the node embeddings and the attention mechanism.

To capture the invariant and variant patterns, we then devise a disentangling mechanism to separate the evolving structural and featural patterns. Specifically, we first calculate the attention scores among nodes in the dynamic neighborhood to obtain the structural masks,
\begin{equation}
\begin{aligned}
\mathbf{m}^I&=\text{Softmax}(\frac{\mathbf{q}\cdot \mathbf{k}^\top}{\sqrt{d}}),
\mathbf{m}^V=\text{Softmax}(-\frac{\mathbf{q}\cdot \mathbf{k}^\top}{\sqrt{d}}),\\
\end{aligned}
\end{equation}
where $\mathbf{m}^I$ and $\mathbf{m}^V$ represent the masks of invariant and variant structural patterns. 
This approach aims to assign higher attention scores to dynamic neighbors with invariant patterns and lower attention scores to those with variant patterns, resulting in a negative correlation between the two types of patterns. To capture invariant featural patterns, we introduce a learnable featural mask, denoted as $\mathbf{m}_f$, which is obtained by applying a Softmax function to a learnable weight vector $\mathbf{w}_f$. This featural mask allows us to selectively choose features from the messages of dynamic neighbors. By summarizing the messages of the dynamic neighborhood with the respective masks, our approach captures both invariant and variant patterns in a dynamic neighbor relationship,\ie,

\begin{equation}
\centering
\small
\begin{aligned}
&\tilde{\mathbf{z}}_t^I(u) = \sum_i \mathbf{m}^I_i (\mathbf{v}_i\odot \mathbf{m}_f),
\mathbf{z}_t^I(u) =\text{FFN}(\tilde{\mathbf{z}}_t^I(u)+ \mathbf{h}_t(u)),
\\
&\tilde{\mathbf{z}}_t^V(u) = \sum_i \mathbf{m}^V_i  \mathbf{v}_i,
\mathbf{z}_t^V(u) =\text{FFN}(\tilde{\mathbf{z}}_t^V(u)),
\label{eq:emb}
\end{aligned}
\end{equation}
where the FFN includes a layer normalization~\cite{ba2016layer}, multi-layer perceptron and skip connection,\ie, 
\begin{equation}
    \text{FFN}(\mathbf{x})=\alpha \cdot \text{MLP(LayerNorm}(\mathbf{x}))+(1-\alpha) \cdot \mathbf{x},
\end{equation}
where $\alpha$ is a learnable parameter. The pattern summarizations are then combined as hidden embeddings, which are subsequently fed into subsequent layers for further processing,\ie,
\begin{equation}
    \mathbf{h}_t(u) \leftarrow \mathbf{z}_t^I(u)+\mathbf{z}_t^V(u).
\end{equation}

The proposed architecture consists of a stack of disentangled graph attention layers. This design allows each node to indirectly access dynamic neighborhoods of higher order, similar to traditional graph message-passing networks. At the $l$-th layer, the hidden representations $\mathbf{z}_t^I(u)$ and $\mathbf{z}_t^V(u)$ are obtained as summarizations of the invariant and variant patterns, respectively, within the $l$-order dynamic neighborhood. Following~\cite{vaswani2017attention,sankar2020dysat}, we extend the attention to multi-head attention to improve modeling ability and stability.

\subsection{Invariant Dynamic Graph Regression}
The data in real-world applications is complex that may consist of various patterns, where the variant patterns have variant relationship with labels, and if the model relies on these patterns, it will evitably have deteriorated prediction performance under distribution shift. Following OOD generalization literature~\cite{zhang2022dynamic, wu2022discovering}, we propose to reduce the variance of the model's prediction when the model is exposed to the same invariant pattern and different variant patterns. We first approximate the patterns $\mathbf{P}^t$ with summarized patterns $\mathbf{z}_t$ in the previous section. 
Since the hidden representations $\mathbf{z}_t^I(u)$ and $\mathbf{z}_t^V(u)$ serve as summarizations of the invariant and variant patterns for node $u$ at time $t$, we aim to approximate the model's prediction abilities under exposure to various variant patterns. To achieve this, we adopt a sampling-based approach whereby we collect the variant patterns of all nodes at all times. We then randomly sample one variant pattern and use it to replace the variant patterns of other nodes across different time steps. For instance, we can replace the variant pattern of node $u$ at time $t_1$ with the variant pattern of node $v$ at time $t_2$ as follows:
\begin{equation}
    \mathbf{z}_{t_1}^I(u),\mathbf{z}_{t_1}^V(u)\leftarrow \mathbf{z}_{t_1}^I(u),\mathbf{z}_{t_2}^V(v).
    \label{eq:intervene_node}
\end{equation}
By replacing the variant patterns of nodes in this manner, we can approximate the effects of exposure to different variant patterns on the overall dynamics of the graph. As the invariant patterns are assumed to sufficiently determine the labels and the influence of the variant patterns on the labels are shielded given the invariant patterns, the labels should not be changed. Therefore, to let the model focus on invariant patterns to make predictions, we introduce an invariance loss to minimize the variance of the model predictions when exposed to various variant patterns. 
Given the summarized invariant and variant patterns $\mathbf{z}_I, \mathbf{z}_V$, we calculate the task loss and mixed loss as 
\begin{equation}
\mathcal{L}_{\text{task}}=\ell( f(\mathbf{z}^I),\mathbf{y}), \mathcal{L}_{\text{mix}}\mid \mathbf{z}^V =\ell( g(\mathbf{z}^V,\mathbf{z}^I),\mathbf{y}),
\label{eq:loss}
\end{equation}
where $f(\cdot)$ is a MLP regression predictor, and $g(\mathbf{z}^I,\mathbf{z}^V) = \text{MLP}(\mathbf{z}^I + \text{sigmoid}(\mathbf{z}^V))$ makes predictions with variant patterns. The task loss encourages the model to utilize the invariant patterns, while the mixed loss measures the model's prediction ability when variant patterns are also exposed to the model. Then the invariance loss is calculated by 
\begin{equation}
\mathcal{L}_{\text{inv}}=\frac1{|\mathcal{S}|}\sum_{\mathbf{z}^V \in \mathcal{S}}(\mathcal{L}_{\text{mix}}\mid\mathbf{z}^V) + \text{Var}_{\mathbf{z}^V \in \mathcal{S}}(\mathcal{L}_{\text{mix}}\mid\mathbf{z}^V),
\label{eq:inv}
\end{equation}
where $\mathcal{S}$ is the set of variant patterns. The invariance loss measures the variance of the model's prediction under multiple intervened distributions. The final training objective is
\begin{equation}
\label{eq:final}
\min_{\theta} \mathcal{L}_{\text{task}} +  \lambda \mathcal{L}_{\text{inv}},
\end{equation}
where the task loss $\mathcal{L}_{\text{task}}$ is minimized to exploit invariant patterns, while the invariance loss $\mathcal{L}_{\text{inv}}$ help the model to discover invariant and variant patterns, and $\lambda$ is the hyperparameter to balance between two objectives. After training, we only adopt invariant patterns
to make predictions in the inference stage. The overall algorithm is summarized in Algorithm~\ref{algo:pipeline}. The framework is illustrated in Figure~\ref{fig:framework}.

\begin{algorithm}
\caption{Training pipeline for \modelnosp} 
\label{algo:pipeline}
\begin{algorithmic}[1]
\REQUIRE 
Training epochs $L$, the sampling number of variant patterns $S$, the hyperparameter $\lambda$.
\FOR{$l = 1, \dots, L$}
    \STATE Obtain $\mathbf{z}_t^V(u),\mathbf{z}_t^I(u)$ for each node and time as Eq.~\eqref{eq:emb}
    \STATE Sample $S$ variant patterns from the collections of $\mathbf{z}_t^V(u)$ to construct the set of variant patterns $\mathcal{S}$
    \FOR{$s$ in $\mathcal{S}$}
        \STATE Replace the nodes' variant pattern summarizations with $s$ as Eq.~\eqref{eq:intervene_node}
        \STATE Calculate the mixed loss as Eq.~\eqref{eq:loss}
    \ENDFOR
    \STATE Update the model according to Eq.~\eqref{eq:final}
\ENDFOR
\end{algorithmic}
\end{algorithm}

\section{Experiments}
In this section, we conduct extensive experiments to verify the design of our framework. 

\subsection{Baselines}
We adopt several representative statistical models, sequence neural network models, dynamic graph neural network models and out-of-distribution generalization methods. The first group of these methods is statistical models, including:

\begin{itemize}
	\item \textbf{MA} (Moving Average Model)~\cite{durbin1959efficient, box2015time} is a widely used statistical model in time series analysis. It assumes that the current value of a time series is a linear combination of the past error terms, \ie, the moving average residuals.
	
	\item \textbf{ARMA} (Autoregressive Moving Average Model)~\cite{gurland1954hypothesis, box2015time} is a time series forecasting model that combines the Auto-regressive model with the MA model. It assumes that the current value of a time series depends on its previous values and a linear combination of its past errors.

	\item \textbf{ARIMA} (Autoregressive Integrated Moving Average Model)~\cite{box2015time} is a time series forecasting model that combines the ARMA model with the concept of differencing. It assumes the current value of a time series depends on its previous values, a linear combination of its past errors, and the difference between its current and past values.
\end{itemize} 
The second group of these methods is sequence neural network models, including:
\begin{itemize}
	\item \textbf{RNN}~\cite{rumelhart1986learning} is a neural network designed for sequential data processing. It uses a feedback loop to process previous inputs and generate outputs. RNNs are widely used in natural language processing, speech recognition, and time series forecasting.
	
	\item \textbf{GRU}~\cite{chung2014empirical} is a variant of RNN that uses gating mechanisms to control the flow of information. It has fewer parameters than the traditional RNN and is faster to train.
	
	\item \textbf{LSTM}~\cite{hochreiter1997long} is another variant of RNN that uses memory cells to store information for long periods. It can selectively remember or forget information and is effective in various applications such as language modeling and speech recognition.
\end{itemize} 
The third group of these methods is dynamic graph neural network models, including:
\begin{itemize}
	\item \textbf{GRUGCN}~\cite{seo2018structured} is a neural network architecture that combines the GRU~\cite{Cho2014LearningPR} and graph convolutional network (GCN)~\cite{kipf2016variational} modules. It is designed for graph-structured data and has been shown to achieve state-of-the-art performance on various dynamic graph tasks.
	
	\item \textbf{EGCN}~\cite{pareja2020evolvegcn} is another neural network architecture designed for graph-structured data. It adopts an LSTM~\cite{hochreiter1997long} or GRU~\cite{Cho2014LearningPR} to flexibly evolve the GCN~\cite{kipf2016variational} parameters instead of directly learning the temporal node embeddings, which applies to frequent change of the node set on dynamic graphs.
	
	\item \textbf{DySAT}~\cite{sankar2020dysat} is a neural network architecture designed for dynamic graphs. It aggregates neighborhood information at each graph snapshot using structural attention and models network dynamics with temporal self-attention. In this way, the weights can be adaptively assigned for the messages from different neighbors in the aggregation, so that the complex network dynamics in the real-world datasets can be well captured. 
\end{itemize}
The last group of these methods is out-of-distribution (OOD) generalization methods, including:
\begin{itemize}
	\item \textbf{IRM}~\cite{arjovsky2019invariant} is a general method designed to enforce invariance in machine learning models. It aims at learning an invariant predictor which minimizes the empirical risks for all training domains to achieve OOD generalization. 

	\item \textbf{VREx}~\cite{krueger2021out} is an optimization method designed for training deep learning models. To improve the out-of-distribution generalization abilities of the model, it reduces the differences in the risks across training domains to decrease the model’s sensitivity to distributional shifts.

	\item \textbf{GroupDRO}~\cite{sagawa2019distributionally} is another optimization method designed to improve the robustness of deep learning models to distributional shifts. It puts more weight on training domains with larger errors when minimizing empirical risk to minimize worst-group risks across training domains.
\end{itemize}

As the out-of-distribution generalization methods are general methods for machine learning models, and they are not specially designed for human albumin prediction, we adopt the best-performed baseline DySAT as their backbones.

\subsection{Experimental Setups}
In this section, we introduce the experimental setups, including data preprocessing, edge construction, and training and evaluation protocols.

\subsubsection{Data Preprocessing} The original data have various categorical features and continuous features with different scales. For example, in training data, the attribute MCH has values ranging from 24.9 to 34.9, while the values of the attribute ALT can range from 4.9 to 11272.2. To make the data suitable for the model, we first convert the category features into one-hot vectors and then normalize the continuous features by a standard scaler,\ie, $x' = \frac{x - \mu}{\sigma}$, where $\mu$ and $\sigma$ are the mean and standard deviation of the feature values, respectively. Note that we only adopt the training data to calculate the mean and standard deviation of the features to avoid information leakage. 

\subsubsection{Edge Construction} Patients with similar physiological characteristics may have similar albumin curves, and the graph neural network (GNN) models can leverage this information to improve the albumin prediction accuracy. However, the original data only contain the evolving features of the patients during hospitalization, without considering the relationship among them. Therefore, we construct the relationship between patients by measuring the similarity of the patient physiological characteristics as follows. We first sort the data by the time of the data and then for each time we construct a graph with patients as nodes, and the relationship between patients as edges. We normalize the physiological characteristics of the patients by a standard scaler, calculate the L1 distance between the patients, and then adopt a K-nearest-neighbors (KNN)~\cite{cover1967nearest, fix1989discriminatory} algorithm to construct the edges for each node. We set K=100 in our experiments. In this way, a dynamic graph is constructed based on the physiological characteristics of the patients each day during hospitalization, which can capture the relationship between patients evolving through time.

\subsubsection{Training and Evaluation} For neural network models, we adopt the Adam optimizer~\cite{kingma2014adam} to train the models with a learning rate of 1e-2 and a weight decay of 5e-7. We train the models for a maximum epoch of 1000 and use the early stopping strategy to avoid overfitting. We set the patience to 50, \ie, if the validation loss does not decrease for 50 epochs, we stop the training process. We set the hidden dimensionality for model parameters as 8 and the dimensionality for category attributes as 2. We use the mean squared error (MSE) as the loss function to train the models, \ie, $\ell(\hat y,y) = (\hat y - y)^2$, where $\hat y$ is the prediction value and $y$ is the label. We calculate the loss for each time and node and aggregate all the loss for updating the model parameters,\ie, $\mathcal{L} = \frac1N \frac1T \sum_t \sum_n \ell(\hat y_{n,t}, y_{n,t})$, where the subscript $n$ denotes the patient index and the subscript $t$ denotes the time index. For predicting the albumin of patient $n$ at time $t$, the model only has the patient features before time $t$. 
We adopt Root Mean Square Error (RMSE) and Mean Absolute Error (MAE) as metrics to evaluate the models. The model with the best MAE metric is adopted for evaluation. We run all the experiments with different seeds and initializations 3 times and report the average results and standard deviations. For a fair comparison, the training and evaluation protocols are kept the same for all methods.

\subsection{Experimental Results}

\begin{table*}[htbp]
\centering
\caption{Results of different methods on the ANIC dataset. The best results are in bold and the second-best results are underlined. The experiments are randomly run with three different initialization, and the  average results and the standard deviations are reported. The results of each disease category are also reported.}
\label{tab:main}
\adjustbox{max width = \textwidth}
{
\begin{tabular}{ccccccccc}
\midrule
Model    & \multicolumn{2}{c}{Trauma Disease}                      & \multicolumn{2}{c}{Surgery Disease}                     & \multicolumn{2}{c}{Internal Disease}                    & \multicolumn{2}{c}{Average}                     \\
Metric   & RMSE                   & MAE                    & RMSE                   & MAE                    & RMSE                   & MAE                    & RMSE                   & MAE                    \\ \midrule
MA       & \ms{3.3528}{0.0000}    & \ms{2.7895}{0.0000}    & \ms{4.0516}{0.0000}    & \ms{3.5470}{0.0000}    & \ms{3.8950}{0.0000}    & \ms{3.0374}{0.0000}    & \ms{3.8604}{0.0000}    & \ms{3.0319}{0.0000}    \\
ARMA     & \ms{3.6149}{0.0000}    & \ms{2.9261}{0.0000}    & \ms{4.6298}{0.0000}    & \ms{4.1769}{0.0000}    & \ms{4.1861}{0.0000}    & \ms{3.3159}{0.0000}    & \ms{4.1781}{0.0000}    & \ms{3.3140}{0.0000}    \\
ARIMA    & \ms{3.2952}{0.0000}    & \ms{2.7156}{0.0000}    & \ms{4.3269}{0.0000}    & \ms{3.7018}{0.0000}    & \ms{3.6535}{0.0000}    & \ms{2.7750}{0.0000}    & \ms{3.7501}{0.0000}    & \ms{2.8656}{0.0000}    \\
RNN      & \ms{2.7209}{0.1172}    & \ms{2.0584}{0.1141}    & \ms{3.2646}{0.1053}    & \ms{2.8736}{0.0762}    & \ms{3.3744}{0.1114}    & \ms{2.6699}{0.0784}    & \ms{3.2675}{0.1018}    & \ms{2.5359}{0.0618}    \\
GRU      & \ms{2.8136}{0.1299}    & \ms{2.1774}{0.1836}    & \ms{3.4725}{0.2003}    & \ms{2.9776}{0.1938}    & \ms{3.3329}{0.1046}    & \ms{2.6048}{0.1063}    & \ms{3.2846}{0.0577}    & \ms{2.5375}{0.0791}    \\
LSTM     & \ms{2.7848}{0.1031}    & \ms{2.0612}{0.0214}    & \ms{3.3591}{0.0559}    & \ms{2.9079}{0.0562}    & \ms{3.5191}{0.3712}    & \ms{2.7234}{0.2413}    & \ms{3.3952}{0.2226}    & \ms{2.5742}{0.1504}    \\
GRUGCN   & \ms{3.9567}{0.9721}    & \ms{3.2579}{0.8606}    & \ms{5.0171}{1.0362}    & \ms{4.4816}{0.8725}    & \ms{5.0763}{0.1233}    & \ms{4.0856}{0.1398}    & \ms{4.8801}{0.4004}    & \ms{3.9180}{0.4078}    \\
EGCN     & \ms{4.1087}{0.9669}    & \ms{3.3941}{0.8116}    & \ms{4.8549}{0.8622}    & \ms{4.2420}{0.7816}    & \ms{4.9073}{0.2594}    & \ms{4.0160}{0.2443}    & \ms{4.7642}{0.4882}    & \ms{3.8819}{0.4512}    \\
DySAT    & \ms{2.6650}{0.0077}    & \ms{2.0425}{0.0217}    & \ms{3.2007}{0.0359}    & \ms{2.8256}{0.0073}    & \ms{3.2763}{0.0476}    & \ms{2.5900}{0.0342}    & \ms{3.1838}{0.0303}    & \ms{2.4761}{0.0164}    \\
IRM      & \msone{2.6231}{0.0730} & \mstwo{2.0118}{0.0792} & \msone{3.1864}{0.0353} & \msone{2.7711}{0.0437} & \ms{3.2630}{0.0440}    & \ms{2.5800}{0.0490}    & \ms{3.1595}{0.0378}    & \ms{2.4557}{0.0502}    \\
VREx     & \ms{2.6651}{0.0077}    & \ms{2.0427}{0.0220}    & \ms{3.2009}{0.0356}    & \ms{2.8258}{0.0073}    & \ms{3.2757}{0.0485}    & \ms{2.5896}{0.0347}    & \ms{3.1835}{0.0308}    & \ms{2.4759}{0.0166}    \\
GroupDRO & \ms{2.6585}{0.0286}    & \ms{2.0347}{0.0361}    & \mstwo{3.1876}{0.0304} & \mstwo{2.7791}{0.0365} & \mstwo{3.2142}{0.0554} & \mstwo{2.5422}{0.0680} & \mstwo{3.1429}{0.0398} & \mstwo{2.4388}{0.0500} \\ \midrule
\model   & \mstwo{2.6330}{0.0129} & \msone{2.0034}{0.0033} & \ms{3.2469}{0.0312}    & \ms{2.8916}{0.0421}    & \msone{3.1737}{0.0746} & \msone{2.4919}{0.0539} & \msone{3.1175}{0.0532} & \msone{2.4119}{0.0357} \\ \bottomrule
\end{tabular}
}
\end{table*}

\begin{figure*}
\centering
\includegraphics[width = 0.95\textwidth]{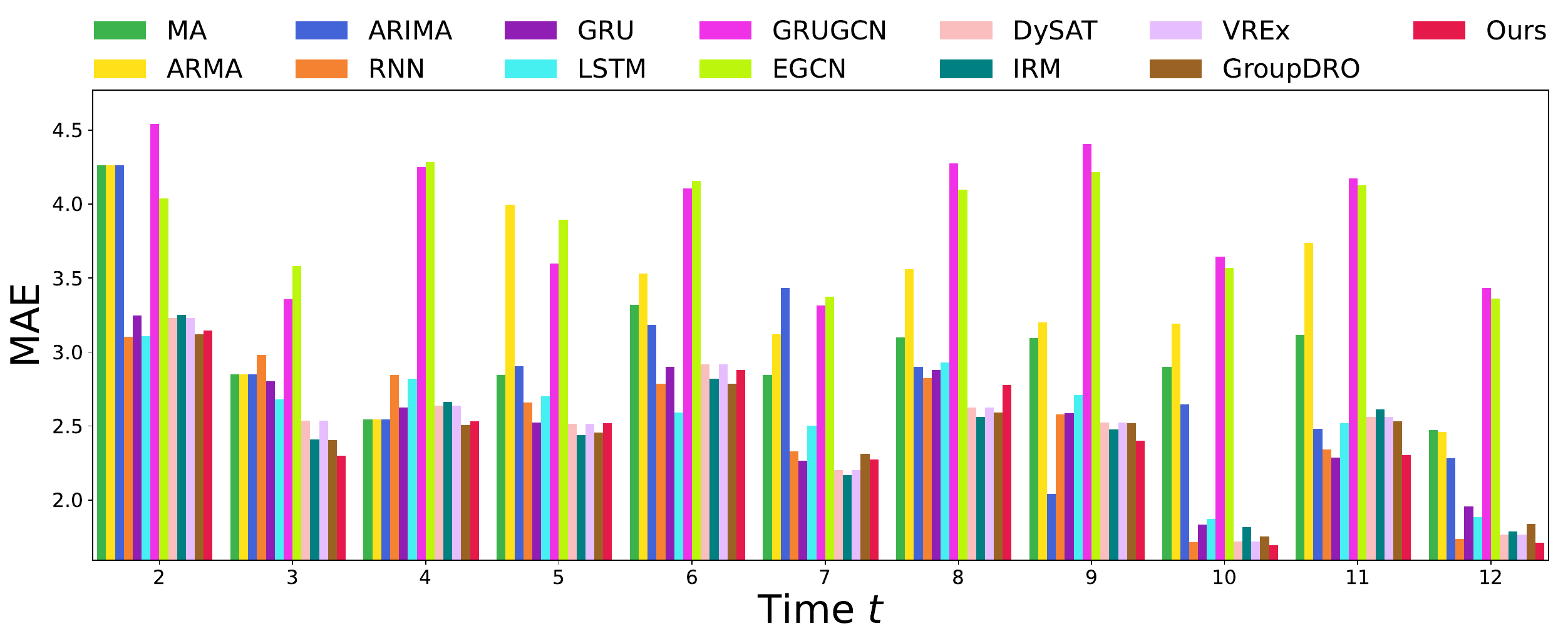}
\caption{Comparisons of different methods in terms of mean absolute error (MAE) at each time.}
\label{fig:time_mae}
\end{figure*}

\subsubsection{Main Results}
In this experimental analysis, we compare the performance of different methods for predicting patient albumin. As shown in Table~\ref{tab:main}, we have the following observations: 
\begin{itemize}
    \item Sequence neural network models outperform statistical models, due to their ability to capture the relationships between multiple variables and assist in albumin trend prediction.
    \item Dynamic graph neural networks are effective in further leveraging the similarities between patients by message passing and information aggregation, strongly improving the albumin prediction accuracy. An exception is that GRUGCN and EGCN which both use GCN for structural modeling, may suffer from over-smoothing problems, as they can not assign different weights to different messages like DySAT does using attention mechanisms.
    \item General OOD generalization methods improve the performance of the best-performed backbone DySAT across all categories, which highlights the existence of distribution shifts in real-world data and verifies the effectiveness of OOD generalization methods.
    \item Our method achieves significantly better results in terms of average RMSE and MAE over all the baselines. This can credit to our method's capability to handle both the dynamic changes of features and structures and identify the invariant patterns to resist distribution shifts, enabling reliable performance even with more complex modeling. 
\end{itemize}
Figure~\ref{fig:time_mae} demonstrates the MAE of different methods on different days, which also shows that our method has significantly better performance of albumin prediction on most days.

\subsubsection{Showcases}
\begin{figure*}
\centering
\includegraphics[width = 0.95\textwidth]{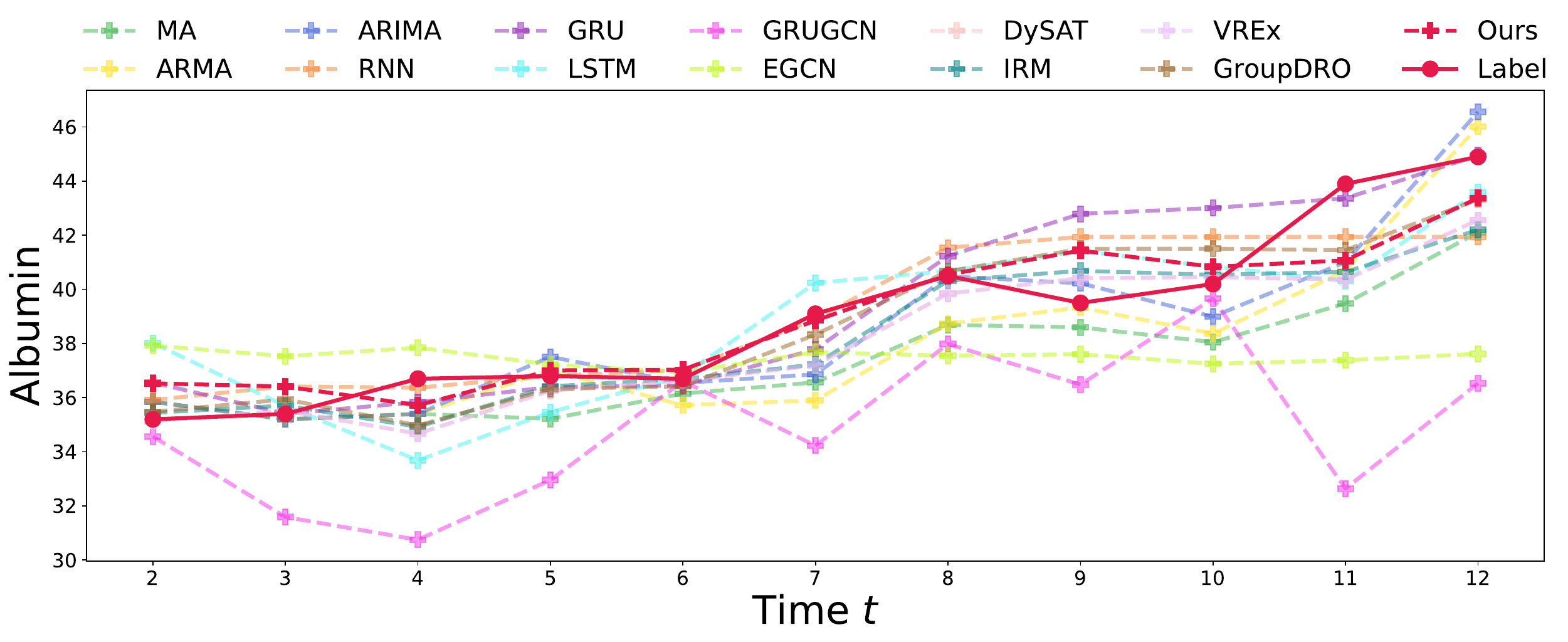}
\caption{A showcase of the albumin predictions of different methods.}
\label{fig:case1}
\end{figure*}

We showcase the albumin prediction of different methods in 
Figure~\ref{fig:case1}. We can find that our method is able to accurately predict the patient albumin value as well as its trend the next day, which verifies the design of our method in the capability of exploiting the feature and structure dynamics in real-world data.

\subsubsection{Feature Importance Analyses}
In this section, we utilize our method to calculate and analyze the importance of various features in the data. Similar to~\cite{selvaraju2017grad}, we measure the feature importance by calculating the derivative of the loss with respect to the features, \ie, $\text{Imp}(x,t) = \frac1N \sum_{n}|\partial{\mathcal{L}_{n,t}}/\partial{x}|$, where $\mathcal{L}_{n,t}$ is the loss of the $n$-th patient at time $t$, and $N$ is the number of patients. For categorical variables, we calculate the gradients of the loss with respect to its corresponding embeddings, and for a variable with multiple dimensions, we calculate the average importance of all the dimensions as the importance of this categorical feature.

\begin{figure*}
\centering
\includegraphics[width = 0.95\textwidth]{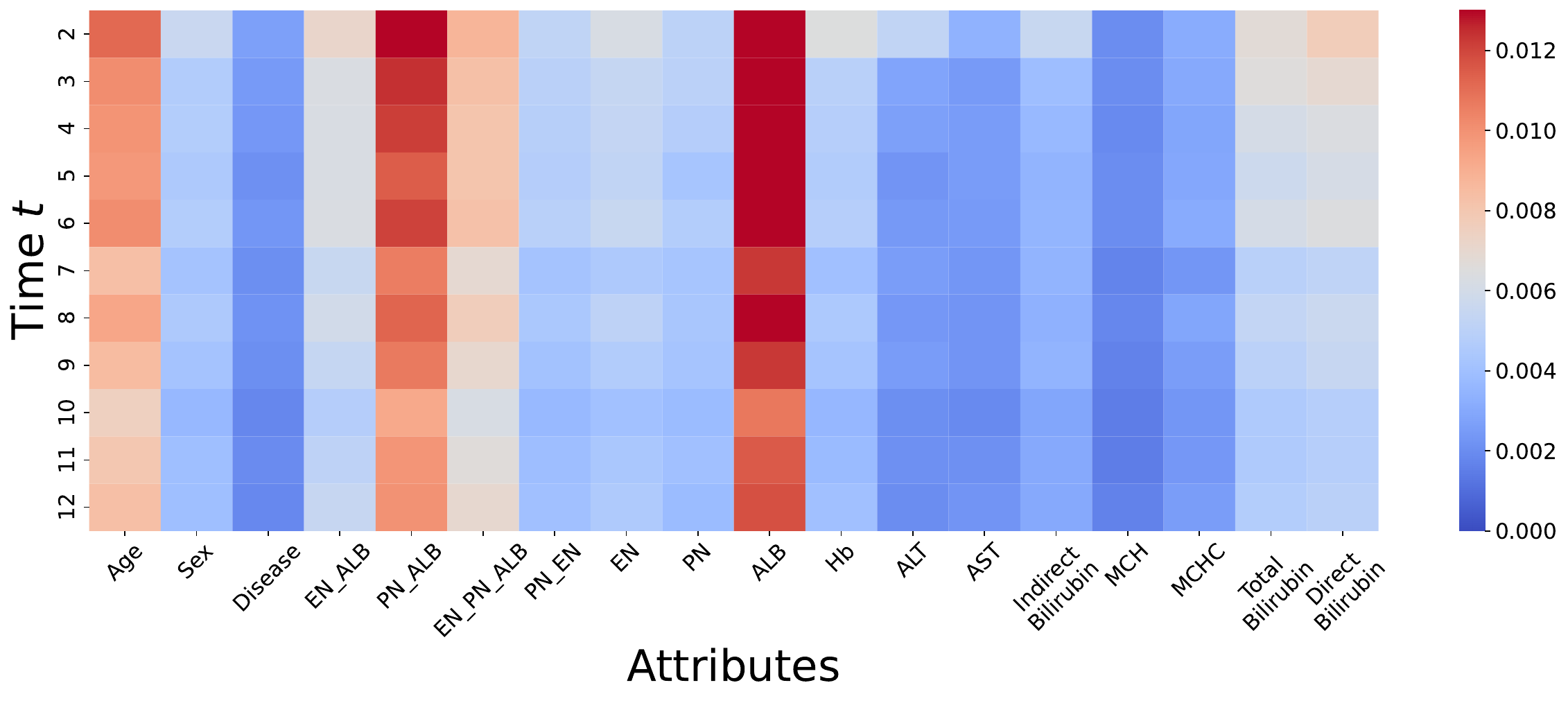}
\caption{The importance of patient features given by our model.}
\label{fig:explain}
\end{figure*}

 We consider the average length of stay in the intensive care unit, typically around 12 days\cite{baker2019characterisation}. Consequently, we select the monitoring data of a patient hospitalized for 12 consecutive days. Figure~\ref{fig:explain} illustrates the visualization results, from which several key observations can be made. 
 The four factors deemed important in the prediction of albumin levels align with common medical knowledge: ALB, EN(\_PN)\_ALB, age, and bilirubin. ALB and EN(\_PN)\_ALB plays a significant role in elevating blood albumin levels by directly supplementing patients with human albumin, either orally or intravenously. Additionally, as the body's metabolic capacity naturally declines with age, it affects albumin's synthesis and breakdown processes. Elevated bilirubin levels are linked to impaired liver function and biliary tract issues, which can impact the synthesis and stability of albumin, thereby influencing blood albumin levels\cite{quinlan2005albumin}.

The experimental results underscore the importance of ALB, EN(\_PN)\_ALB, age, and bilirubin in predicting albumin levels. These indicators reflect the body's nutritional status, liver function, and digestive and absorption capacity, which are crucial in maintaining normal albumin levels.

\subsection{Configurations}

We implement our method with PyTorch, and conduct the experiments with:
\begin{itemize}[leftmargin=0.5cm]
    \item \red{Operating System: Ubuntu 18.04.1 LTS}
    \item \red{CPU: Intel(R) Xeon(R) Gold 6240R CPU @ 2.40GHz}
    \item \red{GPU: NVIDIA GeForce RTX 3090 with 24 GB of memory}
    \item \red{Software: Python 3.8.13, Cuda 11.7, PyTorch~\cite{paszke2019pytorch} 2.0.0, PyTorch Geometric~\cite{Fey/Lenssen/2019} 2.0.3.}
\end{itemize}

\section{Discussions}
\subsection{Limitations of Our Work}
While our research showed promising results in utilizing dynamic GNNs for albumin prediction, there are several limitations that should be acknowledged. Firstly, our study focuses primarily on the analysis of distribution shift, time series patterns, and complex patient relationships. However, there might be other relevant factors influencing albumin levels, such as genetic markers or medical interventions, which are not fully explored in this work. Incorporating additional features and contextual information could potentially enhance the predictive performance of our model.

\subsection{Future Directions}
There are several directions for future research in the field of albumin prediction. Firstly, incorporating additional data sources, such as electronic health records or genetic information, could provide a more comprehensive understanding of the factors influencing albumin levels. This integration of diverse data modalities has the potential to improve the model's predictive accuracy\cite{li2021jlan}.
Furthermore, investigating the interpretability of dynamic GNNs in the context of albumin prediction is crucial for gaining trust and acceptance from healthcare professionals.
Collaborations with medical professionals and domain experts will be invaluable in guiding the development of albumin prediction models. 

By addressing these challenges and incorporating advancements from interdisciplinary fields, dynamic GNNs have the potential to revolutionize albumin prediction and contribute to personalized healthcare.

\section{Related works}
\subsection{Deep learning for Medicine}

Deep learning has made remarkable progress in medical image analysis\cite{chan2020deep}, particularly in medical image recognition, classification, and radiology report generation\cite{rehman2021survey}. Traditional methods in this field rely on manually designed feature extraction and classification algorithms, which face challenges when dealing with complex structures and evolving patterns. Shin\etal~\cite{shin2016learning} employ a CNN-RNN architecture to identify diseases from chest X-ray images based on visual attributes. Nevertheless, medical report generation aims to produce comprehensive reports rather than isolated sentences. Li\etal~\cite{li2018hybrid} employ a reinforcement learning approach to update neural networks by incorporating sentence-level and word-level incentives, while simultaneously ensuring high diagnostic accuracy in the generated reports. Chen\etal~ \cite{DBLP:conf/emnlp/ChenSCW20} propose a memory-based transformer model to investigate the correlation between medical images.

Similarly, deep learning plays a crucial role in clinical decision support by leveraging medical data, including patients' clinical characteristics and medical literature. Instance-based recommendation models\cite{hoang2019learning,shang2019gamenet} focus on the patient's health status for treatment advice, while longitudinal-based models\cite{li2023dgcl} incorporate longitudinal patient history and capture time dependence in recommendations. Li\etal~\cite{li2022knowledge} propose the KDGN model that utilizes a bipartite graph coding structure to mine potential therapeutic agents. Bhoi\etal~\cite{bhoi2021personalizing} work on personalized drug recommendations, using GNNs to control the level of drug-drug interactions.

Furthermore, deep learning exhibits promising potential in drug repositioning\cite{ezzat2019computational}. Several computational models have been developed to identify new applications for existing drugs efficiently\cite{lotfi2018review}. For example, Gottlieb\etal~\cite{gottlieb2011predict} utilize the similarity between drugs and diseases to infer potential drug indications. Wang\etal~\cite{wang2013drug} introduce the PreDR model, which leverages heterogeneous information networks to analyze features of drugs. In summary, deep learning has emerged as a powerful tool for medical research and clinical practice. 

\subsection{Dynamic Graph Neural Network}

Dynamic graph neural networks (DyGNNs) have been extensively studied to handle the complex structural and temporal information in dynamic graphs~\cite{skarding2021foundations,zhu2022learnable}. In comparison to Graph neural networks (GNNs)~\cite{zhou2020graph,wang2019heterogeneous,wang2017community,xu2018powerful,wu2020comprehensive}, DyGNNs have to further consider the inherent temporal dimension on dynamic graphs. Some works first tackle the structures and then model the dynamics~\cite{yang2021discrete,sun2021hyperbolic,hajiramezanali2019variational,seo2018structured, sankar2020dysat}, while some others first tackle the temporal information and adopt memory modules to model the structures \cite{wang2021inductive,cong2021dynamic,xu2020inductive,rossi2020temporal}. DyGNNs have been applied in various real-world applications, such as event forecasting~\cite{deng2020dynamic},  dynamic anomaly detection\cite{cai2021structural}, temporal knowledge graph completion~\cite{wu2020temp}, {\it etc}. Some other works automate the GNN designs to adapt to various scenarios~\cite{zhang2023dynamic,qin2022bench,guan2021autogl,qin2021graph,qin2022graph,zhang2023unsupervised,cai2022multimodal,zhou2022curriculum,guan2021autoattend}. To the best of knowledge, this is the first work to solve human albumin prediction problems with dynamic graph neural networks. 

\subsection{Out-of-Distribution Generalization}
In many real-world scenarios, uncontrollable distribution shifts between the training and testing data distributions can occur and result in a significant drop in model performance. To overcome this problem, the Out-of-Distribution (OOD) generalization problem has emerged as a major research topic in various fields~\cite{shen2021towards,yao2022improving}. For the most related to our topic, some recent works have attempted to address distribution shifts on graphs~\cite{li2022out,wu2022discovering,wu2022handling,chen2022invariance,zhu2021shift,qin2022graph,zhang2021revisiting,li2022ood,zhang2022learning,fan2021generalizing,li2022gil}, while some others handle distribution shifts in time-series data~\cite{gagnon2022woods,du2021adarnn,kim2021reversible,venkateswaran2021environment,lu2021diversify,yao2022wildtime}. However, the distribution shifts existing in medical data, especially with dynamic features and structures, are under-explored. In this paper, we study the problem of human albumin prediction under distribution shifts with evolving features and structures.  

\section{Ethical Considerations}
In our study of predicting albumin levels using ICU patients’ data, we prioritize ethical concerns surrounding data privacy and prediction accuracy. All patient data is anonymized, aligning with standards to ensure confidentiality. We are conscious of the ramifications of incorrect predictions, which can impact clinical decisions and patient outcomes. Rigorous validation mechanisms are embedded to enhance the model’s reliability.
The ethical handling of data encompasses informed consent, ensuring patients are aware and agreeable to the use of their anonymized data. 
In sum, our ethical approach ensures data privacy, informed consent, and strives for prediction accuracy, ensuring the welfare of patients is at the core of our research, balancing innovation with ethical integrity.

\section{Conclusions}

In this paper, we proposed a framework named Out-of-Distribution Generalized Dynamic Graph Neural Network for Human Albumin Prediction (\modelnosp) to address the challenges of predicting human albumin levels. We demonstrated the effectiveness of our framework in accurately predicting albumin levels for ICU patients during hospitalization using the proposed dataset ANIC. Through extensive experiments, we showed that our method outperformed state-of-the-art baselines, showcasing its superior performance in human albumin prediction. Our contributions include the novel application of dynamic graph neural networks for this prediction task, the development of the \model framework, the introduction of the ANIC dataset, and the validation of our method's superior performance. These findings highlight the potential of our framework to aid in clinical decision-making and improve patient care by providing accurate predictions of human albumin levels.

\section*{Acknowledgment}
This work was supported in part by the National Key Research and Development Program of China No. 2020AAA0106300, National Natural Science Foundation of China (No. 62250008, 62222209, 62102222, 61936011), Beijing National Research Center for Information Science and Technology under Grant No. BNR2023RC01003, BNR2023TD03006, and Beijing Key Lab of Networked Multimedia. All opinions, findings, conclusions and recommendations in this paper are those of the authors and do not necessarily reflect the views of the funding agencies. 

\bibliographystyle{IEEEtran}
\small
\bibliography{main}
\end{document}